\documentclass[gmd, manuscript]{copernicus}

\begin{document}

\title{Can AI be enabled to dynamical downscaling? A Latent Diffusion Model to mimic km-scale COSMO5.0\_CLM9 simulations}


\Author[1][eltomasi@fbk.eu]{Elena}{Tomasi} 
\Author[1]{Gabriele}{Franch}
\Author[1]{Marco}{Cristoforetti}

\affil[1]{Data Science for Industry and Physics, Fondazione Bruno Kessler, via Sommarive 18, 38123, Trento (TN), Italy}




\runningtitle{Can AI be enabled to dynamical downscaling?}

\runningauthor{Elena Tomasi et al.}

\received{}
\pubdiscuss{} 
\revised{}
\accepted{}
\published{}


\firstpage{1}

\maketitle

\begin{abstract}
Downscaling techniques are one of the most prominent applications of Deep Learning (DL) in Earth System Modeling. A robust DL downscaling model can generate high-resolution fields from coarse-scale numerical model simulations, saving the timely and resourceful applications of regional/local models. Additionally, generative DL models have the potential to provide uncertainty information, by generating ensemble-like scenario pools, a task that is computationally prohibitive for traditional numerical simulations. In this study, we apply a Latent Diffusion Model (LDM) to downscale ERA5 data over Italy up to a resolution of 2 km. The high-resolution target data consists of 2-m temperature and 10-m horizontal wind components from a dynamical downscaling performed with COSMO\_CLM. Our goal is to demonstrate that recent advancements in generative modeling enable DL to deliver results comparable to those of numerical dynamical models, given the same input data, preserving the realism of fine-scale features and flow characteristics. A selection of predictors from ERA5 is used as input to the LDM, and a residual approach against a reference UNET is leveraged in applying the LDM. The performance of the generative LDM is compared with reference baselines of increasing complexity: quadratic interpolation of ERA5, a UNET, and a Generative Adversarial Network (GAN) built on the same reference UNET. Results highlight the improvements introduced by the LDM architecture and the residual approach over these baselines. The models are evaluated on a yearly test dataset, assessing the models' performance through deterministic metrics, spatial distribution of errors, and reconstruction of frequency and power spectra distributions. 
\end{abstract}


\introduction  
High-resolution near-surface meteorological fields such as 2-m temperature and 10-m wind speed are key targets for the weather and climate scientific and operational communities. Such high-resolution information is of essential importance for a wide variety of applications (e.g. available wind potential, predicted energy consumption, etc.), across diverse timescales, from weather forecasting (nowcasting, medium-range forecasting, and seasonal predictions) to climate projections. The hunger for high-resolution data is directly linked to and justified by the information that such data hold: extreme weather events and localized phenomena can typically be described by highly resolved fields only.

Downscaling is a well-known approach that allows obtaining local high-resolution data (predictands) starting from low-resolution information (predictors) by applying suitable refinement techniques. The two most traditional approaches are dynamical downscaling and statistical downscaling \citep{downscaling_types, doi:10.1177/030913339702100403, Maraun_Widmann_2018}, applied alternatively depending on the final goal of each specific application.

Traditionally, high-resolution fields are achieved in operational weather forecasting by performing dynamical downscaling of lower-resolution simulations. Examples of this approach are all the Local Area Models (LAMs) run in every operational center worldwide, fed with global circulation models at a low resolution, and producing high-resolution fields for a localized area, typically nationwide \citep[e.g.,][]{COSMO-DE, AROME, COSMO-ARPAE}. As for the climate community, this approach materializes, for example, in applications run within the WCRP Coordinated Regional Downscaling Experiment (CORDEX, \citet{giorgi58co}), performing dynamical downscaling of climate projections with Regional Climate Models (RCM) going from the $\sim$100-km resolution of Global Climate Models (GCM) down to a $\sim$16-km resolution \citep[e.g.,][and others]{EURO-CORDEX}. The dynamical downscaling approach is well-established and provides physically and temporally consistent fields. However, it still has significant drawbacks due to the high resource demands required for its execution that limit its application, e.g. to deterministic runs instead of (or limited to small) ensemble runs. 

On the other hand, statistical downscaling uses coarse data from numerical simulations to infer data at high resolution by applying empirical relationships or transfer functions derived from a set of known predictors–predictands data pairs \citep{Maraun_Widmann_2018}. Statistical downscaling methods have evolved over the years since the 1990s, with increasingly greater levels of complexity and data. Canonical examples of statistical downscaling methods are linear or multilinear regression methods \citep[e.g.,][]{linearDownScaling, MLR_downscaling}, analog ensemble downscaling \citep[e.g.,][]{AnEns_downscaling} or quantile mapping \citep{panofsky1968some}.

In recent years, the advent of machine-learning techniques introduced many other powerful methods. These approaches have the potential to outperform classical statistical models, introducing nonlinear components in the downscaling process and learning from the provided high-resolution data. Specifically, Convolutional Neural Networks (CNNs) are particularly well suited for handling spatially distributed data and for the super-resolution task, being able to capture complex, nonlinear mappings identifying crucial features, and have already been successfully applied to weather downscaling \citep[e.g.,][]{BanoMedina2020, Rampal2022, Hohlein2020}. Building on CNN frameworks, two Deep Learning (DL) approaches are currently the most promising for improving atmospheric downscaling: Generative Adversarial Networks (GANs, \citet{goodfellow2014generative,goodfellow2020generative}) and Diffusion Models \citep{DiffMod_2015}, which both allow for a probabilistic approach to the problem. The potential and drawbacks of these approaches are reported in the following Section \ref{sec:rel_work_anc_contr}.

In this work, we focus on and test a Latent Diffusion Model (LDM) that is novel for application to the atmospheric downscaling task. Its advantages should be twofold: the diffusion approach has a much more stable training than that of GAN models, still holding the ability to generate small-scale features and the potential for ensemble production, and shows superior results in applications to image processing compared to GANs \citep{Saharia2023, Diff_beats_GAN}. In addition, the latent approach should be seen as incremental compared to pixel-space-based diffusion approaches as it provides a cheaper solution in terms of computational costs for both inference and training \citep{StableDiffusion_2021}. This feature is intrinsically appealing to scale the downscaling task to wider (and longer) domains. Lastly, we use the high-resolution output from a numerical model dynamical downscaling simulation as our target-reference-truth. In doing so, we aim to determine whether a properly trained deep learning model can effectively emulate dynamical downscaling. If successful, this model could serve as a versatile dynamical downscaling emulator, much faster than the traditional numerical dynamical models, with a broad range of utmost significant applications.

\section{Related work and Contribution}
\label{sec:rel_work_anc_contr}
As mentioned above, currently, the most promising DL approaches for improving atmospheric downscaling are Generative Adversarial Networks (GANs) and Diffusion Models, which both are based on CNN frameworks, allow for the generation of small-scale features and for a probabilistic approach to the problem. GANs have already shown promising results in downscaling different meteorological variables, in different regions. For example, \citet{Leinonen2021} applied GANs for reconstructing high-resolution precipitation patterns from coarsened radar images; \citet{Stengel2020} demonstrated GANs potential in performing downscaling of GCMs up to 2km for solar radiation and wind; steps forward pure super-resolution applications have also been made with GANs, as shown in \citet{Harris2022} as well as in \citet{price2022}, where additional variables from numerical models are used as input predictors variables to produce high-resolution precipitation fields. Nevertheless, GANs still pose relevant challenges, such as model instabilities, and mode collapses during the training procedure \citep{arjovsky2017towards, mescheder2018training}.

On the other hand, Diffusion models introduce a relatively younger approach but have already been proven very effective in weather forecasting and nowcasting applications \citep[e.g.,][]{Leinonen2023, Li2024}. Diffusion models have yet to be widely tested and evaluated on the atmospheric downscaling task, but their characteristics and capabilities are undoubtedly promising for this application as shown for example in \citet{Addison2022} and \citet{Mardani2023}.

Building on these encouraging results, in this work, we approach the downscaling task with a Latent diffusion model, comparing it against some standard baselines and a GAN baseline. Specifically, we re-adapted the Latent Diffusion Cast (LDCast) model \citep{Leinonen2023}, recently developed for precipitation nowcasting. LDCast has shown superior performance in the generation of highly realistic precipitation forecast ensembles, and in the representation of uncertainty, compared to traditional GAN-based methods. Our resulting model for downscaling, similar to fully convolutional models, can be trained on examples of smaller spatial domains (patches) and used at the evaluation stage on domains of arbitrary sizes, making it suitable for the generation of high-resolution data covering wider domains. As also suggested in \citet{Mardani2023}, we propose the application of the diffusion model with a residual approach, relying on a standard UNET architecture for capturing the bigger scales and training the latent diffusion model to generate the residual, small scales only. 

Additionally, our work differs from most of the aforementioned works for the chosen pair of low-resolution and high-resolution data for the training. This choice highly influences the level of complexity that the DL downscaling model must achieve. Indeed, downscaling a coarsening of the high-resolution data \citep[e.g.,][]{Leinonen2021, Stengel2020, Vandal2017} is a much easier task than downscaling modeled low-resolution data (e.g. short-term forecasts as in \citet{Harris2022}, seasonal predictions, or climate projections) to independent high-resolution data, either coming from observations or numerical model simulation. While the first exercise falls into a purely super-resolution task, the latter includes learning potential large-scale model biases and correcting them, or detecting and generating local phenomena that cannot be resolved at the coarse resolution of the large-scale models. In our work, we focus on reanalyses products and we train our models using a set of 14 ERA5 variables as low-resolution input and high-resolution data from a dynamical downscaling of ERA5 (run with the COSMO\_CLM model) as target data (2-m temperature and 10-m wind speed horizontal components). This approach is similar to that followed, for example, by \citet{gmd-14-6355-2021} and \citet{Mardani2023}. In doing so, we intentionally force the model to learn to generate the effects of those local phenomena resolved by the dynamical numerical model, emulating its behavior.

\section{Datasets}
\label{sec:datasets}

\subsection{Low- and High-resolution data}
\label{subsec:data}
The goal of this experiment is to train a DL model to mimic a dynamical downscaling performed with a convection-permitting Regional Climate Model (RCM). The target high-resolution data consists of the hourly Italian Very High-Resolution Reanalyses produced with COSMO5.0\_CLM9 (VHR-REA CCLM) by dynamically downscaling ERA5 reanalyses \citep{Hersbach2020} from their native resolution (~25km) up to ~2.2km over Italy \citep{Raffa2021, Adinolfi2023}. Consistently with these target numerical simulations, the input low-resolution data fed to our DL model are ERA5 data.

\subsection{Data alignment and preprocessing}
\label{subsec:datapreproc}
ERA5 data have a resolution of 0.25° worldwide, which roughly corresponds to ~22 km at the latitudes of the focus domain, while VHR-REA CCLM data have a native resolution of 0.02° (2.2 km). Data from both datasets were preprocessed to re-project, trim, and align the low and high-resolution fields. Specifically, the coordinate reference system (CRS) chosen for the experiment is ETRS89-LAEA Europe (Lambert Azimuthal Equal Area), also known in the EPSG Geodetic Parameter Dataset under the identifier EPSG:3035, and the experiment grids align with the European Environmental Agency Reference grid (EEA Reference grid, \citet{EEA_refgrid}). ERA5 was reprojected and interpolated (with nearest-neighbor interpolation) on the EEA 16-km reference grid while VHR-REA CCLM was reprojected and interpolated on the EEA 2-km reference grid. The factor of the downscaling procedure is, therefore, 8x: over the target domain, low-resolution data consists of 72x86 16-km pixel images while high-resolution data consists of 576x672 2-km pixel images.

\subsection{Experimental domain}
\label{subsec:domain}
\begin{figure}
  \centering
  \includegraphics[width=0.6\linewidth]{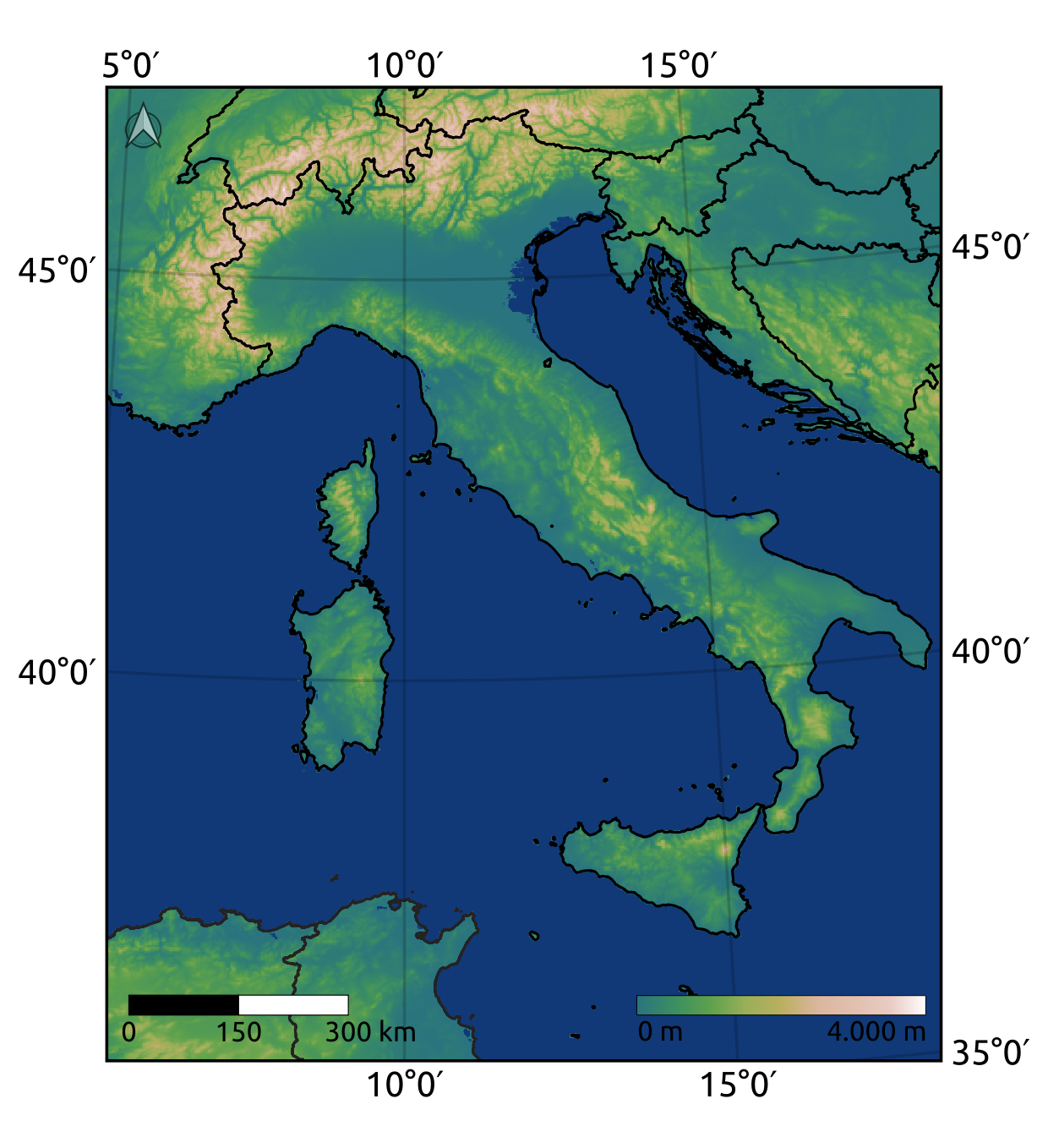}
  \caption{Experimental domain with 2-km digital elevation model.}
  \label{fig:domain}
\end{figure}

The experiment target domain spans from 35\textdegree N to 48\textdegree N, and from 5\textdegree E to 20\textdegree E (Figure \ref{fig:domain}). This area corresponds to the region where VHR-REA CCLM data are available. The region includes a wide variety of topographically different sub-areas (mountainous areas such as the Alps and the Appennini, flat areas such as the Po Valley and coastal lines) which trigger local phenomena whose effects are challenging to identify for the downscaling models, as they are not present in the low-resolution data. 

\subsection{Target variables and predictors}
\label{subsec:datapredictands}
The target variables of the study are (i) 2-m temperature and (ii) horizontal wind components at 10 m: different, dedicated models to each target variable have been trained.
The input ERA5 low-resolution data are both the target variables and additional fields used as dynamical predictors, to improve models' performance. The choice of the set of input fields was based on previous literature \citep[e.g.,][]{Hohlein2020, Rampal2022, Harris2022} and on hardware constraints for the experiment. The selected fields used as predictor variables, for both target variables, are the following, corresponding to a total of 14 input channels to our networks:
\begin{itemize}
\item 2-m temperature
\item 10-m zonal and meridional wind speed
\item mean sea level pressure
\item sea surface temperature
\item snow depth
\item dew-point 2-m temperature
\item incoming surface solar radiation
\item temperature at 850 hPa
\item zonal, meridional and vertical wind speed at 850hPa
\item specific humidity at 850 hPa
\item total precipitation
\end{itemize}

In addition, high-resolution static data have been fed to the models to guide the training and improve performance. These fields include:
\begin{itemize}
\item Digital Elevation Model (DEM)
\item land cover categories
\item latitude
\end{itemize}
DEM data consists of the Copernicus Digital Elevation Model (DEM, \citet{DEM-EU}) interpolated from a resolution of 90 m to a resolution of 2 km. Land cover data were retrieved from the Copernicus Land Service, Global Land Cover data (GLC, \citet{LC_dataset}) interpolated from a resolution of 100 m to a resolution of 2 km. Given that land cover was utilized as a static variable in our analysis, we selected data from 2015: this year represents the earliest epoch available for the selected GLC dataset and falls approximately amid our experimental period of 2000-2020. Land cover class data have been converted to single-channel class masks for the DL models (totaling 16 channels). All static fields have been re-projected and aligned to the high-resolution 2-km EAA reference grid.

\subsection{Dataset splitting strategy}
\label{subsec:datasplitting}
The experiment database comprises hourly data from 2000 to 2020 (~184000 timestamps), for both low and high-resolution data. 70\% of data were used for training, 15\% for validation and 5\% for testing (corresponding to ~15, ~3 and ~1 year, respectively). The testing dataset was fixed to 1 year (5\% of the dataset) because of time constraints in running all the models for the evaluation, especially the diffusion model.

\section{Methods}
\label{sec:methods}
\begin{figure}
    \centering
    \includegraphics[width=0.9\linewidth]{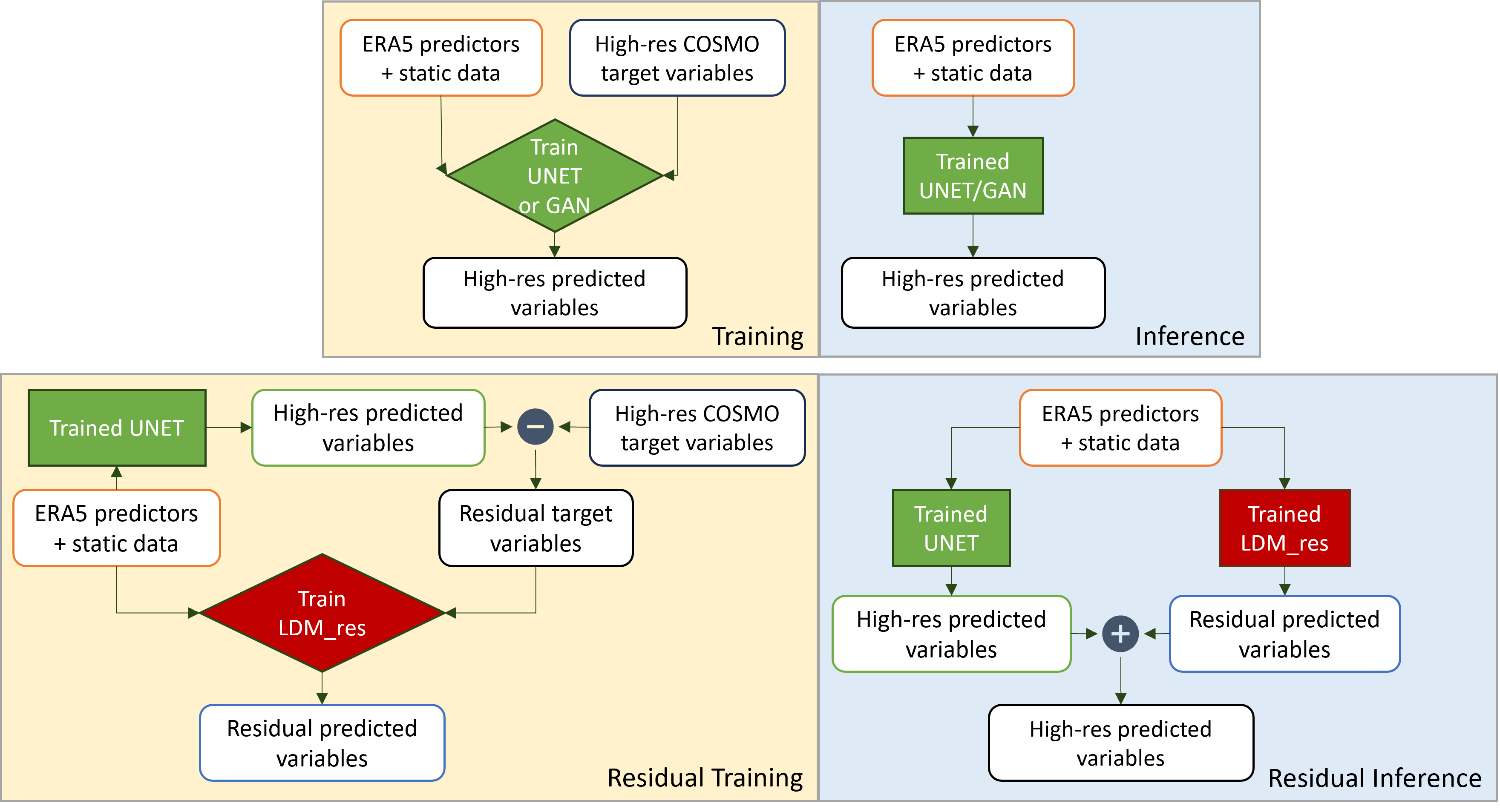}
    \caption{Training and Inference flowcharts for the UNET and GAN models (top row) and for LDM\_res (bottom row): differences between the non-residual and residual approaches are highlighted.}
    \label{fig:ResidualFlowChart}
\end{figure}

In this work, we test a deep generative Latent Diffusion Model (LDM) for the downscaling task, conditioned with low-resolution predictors and high-resolution static data. The implemented LDM is trained to predict the residual error between a previously trained reference UNET and the target variables, hence the model will be addressed as LDM\_res (LDM\_residual) hereafter. This residual approach has shown great performance in the application of pixel-space diffusion models \citep{Mardani2023} and is tested here in the latent diffusion context. The underlying idea is to exploit the great ability of a relatively simple network (a UNET) to properly capture the main, bigger-scale variability of the atmospheric high-resolution data and leverage the power of the generative diffusion model to only focus on the reconstruction of the smaller-scale, locally-driven variability of the fields. Figure \ref{fig:ResidualFlowChart} shows the high-level flow chart of the training and inference setup for LDM\_res, while Section \ref{subsec:modelDiffusion} holds the detailed description of LDM\_res architecture.

LDM\_res is compared against three different baselines with increasing levels of complexity: the quadratic interpolation of ERA5, a UNET, and a Generative Adversarial Network (GAN). The implemented UNET is the core base for each tested deep-learning architecture. Indeed, the same reference UNET network is used (i) as a baseline, (ii) as the Generator of the implemented GAN, and (iii) for the calculation of the residual on which LDM\_res is trained. With this approach, we aim to fairly compare the power of generating small-scale features of the adversarial and the diffusion methods. 

A dedicated network has been trained for each model type, for the two target variables: the 2-m temperature and 10-m horizontal wind components. The downscaling is performed for fixed time steps, with an image-to-image approach.

Given the incremental complexity of the tested models, in the following sections, we start by describing the core reference UNET architecture (Section \ref{subsec:modelUNET}), then the GAN architecture (Section \ref{subsec:modelGAN}), and finally LDM\_res architecture (Section \ref{subsec:modelDiffusion}). Table ~\ref{tab:train_params} shows the number of trainable parameters for each model.

\begin{table}
 \caption{Number of trainable parameters for each tested model.}
  \centering
  \begin{tabular}{llll}
   & & \multicolumn{2}{c}{\textbf{\# of trainable parameters}}\\
    \cline{3-4}
    \textbf{Model}     & \textbf{Sub-net}     & \textbf{per subnet} & \textbf{total}\\
    \hline
    UNET     &    &$\sim$31M    & $\sim$31M  \\
    \hline
    \multirow{2}{*}{GAN} & UNET-Generator  & $\sim$31M     &\multirow{2}{*}{$\sim$34M}\\
     & Discriminator  & $\sim$3M    &\\
     \hline
    \multirow{3}{*}{LDM}    & VAE & $\sim$115K (2mT), $\sim$430K (UV)    &\multirow{3}{*}{$\sim$300M}\\
    & Conditioner  & $\sim$24M     &\\
    & Denoiser  & $\sim$275M     &\\
    \hline
  \end{tabular}
  \label{tab:train_params}
\end{table}

%

\subsection{UNET Architecture}
\label{subsec:modelUNET}
The core UNET network implemented for our experiments is a standard UNET architecture \citep{ronneberger2015unet}, featuring an encoder (contracting path), a bottleneck, and a decoder (expansive path), with skip connections bridging corresponding levels between the encoder and decoder to preserve spatial information. To use a standard UNET to perform downscaling, the input low-resolution data are interpolated with the nearest neighbor interpolation to the target high resolution before feeding them to the network. The encoder is composed of four blocks, each consisting of a layer containing two consecutive 2D convolutions with ReLU activation, interspersed with batch normalization to ensure stable learning \citep{batchNorm}, and a max-pooling operation. The max-pooling layer reduces the spatial resolution by half, enabling the model to capture increasingly complex features while reducing the image dimensions. The output of each encoder block is used in both the next encoder block and the corresponding decoder block through skip connections. The decoder mirrors the encoder with transposed 2D convolutional layers and upsampling steps to go back to the starting resolution. The use of batch normalization ensures robust learning, while the skip connections help preserve critical spatial information across the encoder-decoder bridge. The total number of trainable parameters for the UNET is $\sim$31 M (Table \ref{tab:train_params}). Details on the UNET structure and resolutions are depicted in Figure \ref{fig:UNET_GAN}.

\subsection{GAN Architecture}
\label{subsec:modelGAN}

The Generative Adversarial Network (GAN) \citep{goodfellow2014generative,goodfellow2020generative} tested in this experiment consists of deep, fully convolutional Generator and Discriminator networks, conditioned with low-resolution predictors and high-resolution static data. The generator is trained to output fields that cannot be distinguished from ground truth images by a discriminator, which is trained on the other hand to detect the generator's 'fake' outputs. Our reference GAN consists of a UNET generator upgraded with a Patch-GAN discriminator \citep{Isola_2017_CVPR}. The input data to the generator are low-resolution predictors and high-resolution static data only (no noise addition is performed) and we, therefore, obtain a deterministic GAN.

\begin{figure}
    \centering
    \includegraphics[width=0.7\linewidth]{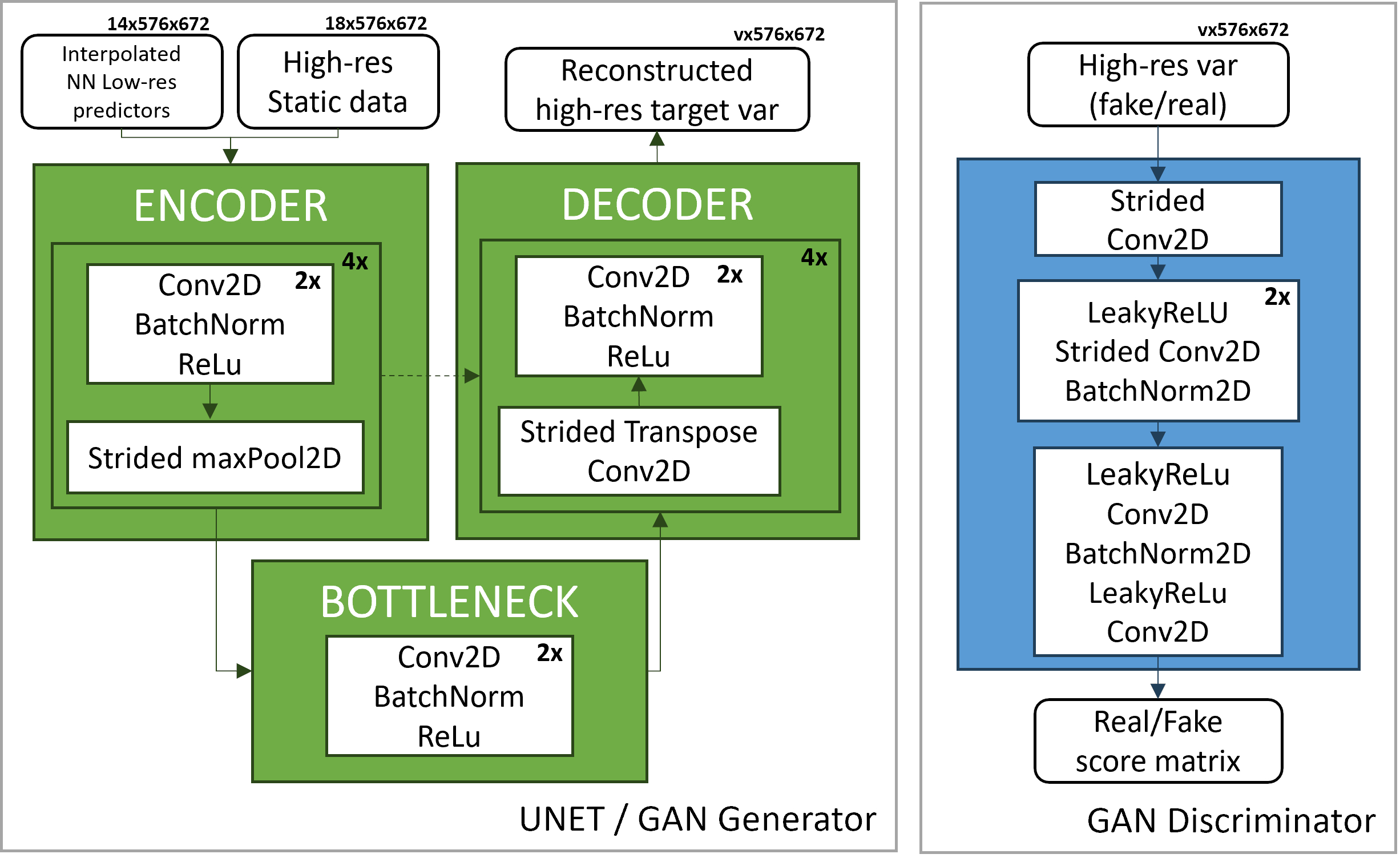}
    \caption{Details on the architectures for the reference UNET and the GAN implemented for the downscaling task. The reference UNET and the generator networks are depicted on the left panel, and the discriminator network is on the right panel. \textit{NN} stands for Nearest Neighbor.}
    \label{fig:UNET_GAN}
\end{figure}

The generator architecture consists exactly of the UNET described in Section \ref{subsec:modelUNET}, as shown in Figure \ref{fig:UNET_GAN}. 

The discriminator is a PatchGAN convolutional classifier \citep{Isola_2017_CVPR}, which focuses on structures at the scale of image patches. The structure of the discriminator is composed of modules of the form convolution-BatchNorm-ReLu. It assigns a "realness" score to each N × N patch of the image, runs convolutionally across the image, and allows us to obtain an overall score by averaging all responses for each patch. High-quality results can be obtained with patches much smaller than the full size of the image, with relevant advantages in terms of resources for the training and application to arbitrarily large images. Details on the discriminator network structure and resolutions are depicted in Figure \ref{fig:UNET_GAN}. The total number of trainable parameters for the GAN is $\sim$34 M (Table \ref{tab:train_params}).

The training procedure follows the combined loss function approach for GANs \citep{goodfellow2020generative}, including recent improvements to promote stability in the training \citep{Esser_2021_CVPR}, with the primary goal of balancing the minimization of both the generator's and discriminator's losses, which are adversarial. The pixel loss we used is the mean absolute error, while the discriminator loss is the hinge loss. The discriminator is activated after 50000 training steps, giving the generator time to learn to generate consistent outputs and thus stabilizing the adversarial training \citep{Esser_2021_CVPR}. After activating the discriminator, the network is trained by updating alternatively the gradients of the generator and the discriminator. The network is constantly fed with the whole target domain and no patch-training is applied.

\subsection{Latent Diffusion Model Architecture}
\label{subsec:modelDiffusion}
Diffusion Models \citep{DiffMod_2015} are probabilistic models meant to extrapolate a data distribution p(x) by corrupting the training data through the successive addition of Gaussian noise (fixed) and then learning to recover the data by reversing this noising process (generative).

The Latent Diffusion Model (LDM) applied for this experiment is an architecture derived from Stable Diffusion \citep{StableDiffusion_2021}, specifically a re-adaptation of the conditional LDM LDCast \citep{Leinonen2023}, developed for precipitation nowcasting and already successfully applied for other variables \citep[e.g.,][]{carpentieri2023extending}. The latent diffusion model derived for the downscaling task is composed of three main elements: (i) a convolutional Variational AutoEncoder (VAE), used to project the residual high-resolution target variables to and from a latent space; (ii) a conditioner, used to embed low-resolution data and high-resolution static data; and a denoiser, used to manage the diffusion process within the latent space. In the following sections, we report a detailed description of each model component. The total number of trainable parameters is shown in Table \ref{tab:train_params}, while Figures \ref{fig:LDM_components} and \ref{fig:LDM_train_inf} summarize the training and inference procedures for the model and the main structure of each component architecture, respectively. 

\subsubsection{Variational Autoencoder}
\label{subsubsec:VAE}
The Variational Autoencoder (VAE) projects the residual high-resolution data from the pixel space to a continuous latent space (encoder) and projects them back to the pixel space (decoder). We train a dedicated VAE for the 2-m temperature and a dedicated VAE for the 10-m wind speed components, independently from the conditioner and denoiser. Once trained, the VAE weights are kept constant during the training of the rest of the network architecture. During inference, only the decoder of the VAE is used (see Figure \ref{fig:LDM_train_inf}).

The encoder and the decoder are structured as 2D convolutional networks composed of blocks of a ResNet residual block \citep{Resnet2008} and a downsampling/upsampling convolutional layer. Three levels of such blocks are used, each reducing each spatial dimension by a factor of 2, while the number of channels is bottlenecked at 32 times the number of input target variables (i.e. 32x1 for the 2-m temperature and 32x2 for the 10-m wind speed components). The VAE bottleneck latent space is regularised with a loss based on Kullback-Leibler divergence (KL) \citep{KL, 10.1214/aop/1176996454} between the latent variable and a multivariate standard normal variable.

While the space dimensions are reduced by a factor of $8^{2}$ (from 512x512 to 64x64 pixel patches), the number of channels is increased from 1 (for the 2-m temperature) and 2 (for the 10-m wind speed components) to 32 and 64, respectively: the overall amount of data is therefore compressed only by a factor of 2, for both the VAEs (from 1x512x512 to 32x64x64 and from 2x512x512 to 64x64x64). Nevertheless, the gain in training performance of the denoiser and conditioner is much greater than the data reduction factor as the compression along the space dimension is more relevant for reducing the computational cost of the training than the increase in channel number \citep{StableDiffusion_2021}.

Figure \ref{fig:LDM_components} depicts details of the VAE's structure.

\subsubsection{Conditioner}
\label{subsubsec:Conditioner}
The conditioner stack acts as a context encoder to process the low-resolution predictors and high-resolution static data and embed them into each level of the denoiser UNET architecture. Initially, both datasets are preprocessed by passing through a dedicated encoder, a projection layer, and an analysis sequence, before being merged.
The predictors' encoder is a basic Identity layer as they already have the same spatial dimensions as the latent space (64x64). The static data encoder is a variational encoder with the same structure as the VAE described in the previous Section \ref{subsubsec:VAE}. For both datasets, the projection layer is a 2D convolutional layer with unitary kernel size, used to increase the number of channels, and the analysis is a sequence of 4 2D Adaptive Fourier Neural Operator (AFNO) blocks (following \citet{pathak2022fourcastnet}), used to extract relevant features. After pre-processing, the conditioning information is prepared to be fed into each level of the denoiser UNET by applying a combination of average pooling and 2D Resnet layers. Figure \ref{fig:LDM_components} depicts details of the conditioner's structure.

\subsubsection{Denoiser}
Our denoising stack is structured as the one of the LDCast \citep{Leinonen2023}, a re-adaptation of the U-Net-type network applied in the original latent diffusion model \citep{StableDiffusion_2021}. The resulting denoiser network consists of a UNET backbone enabled with a conditioning mechanism based on 2D AFNO blocks \citep{Leinonen2023}, aiming at a cross-attention-like operation (as suggested in \citet{guibas2022efficient}). This structure is meant to control the high-resolution synthesis process feeding the conditioning in each level of the UNET architecture.

\begin{figure}
    \centering
    \includegraphics[width=0.9\linewidth]{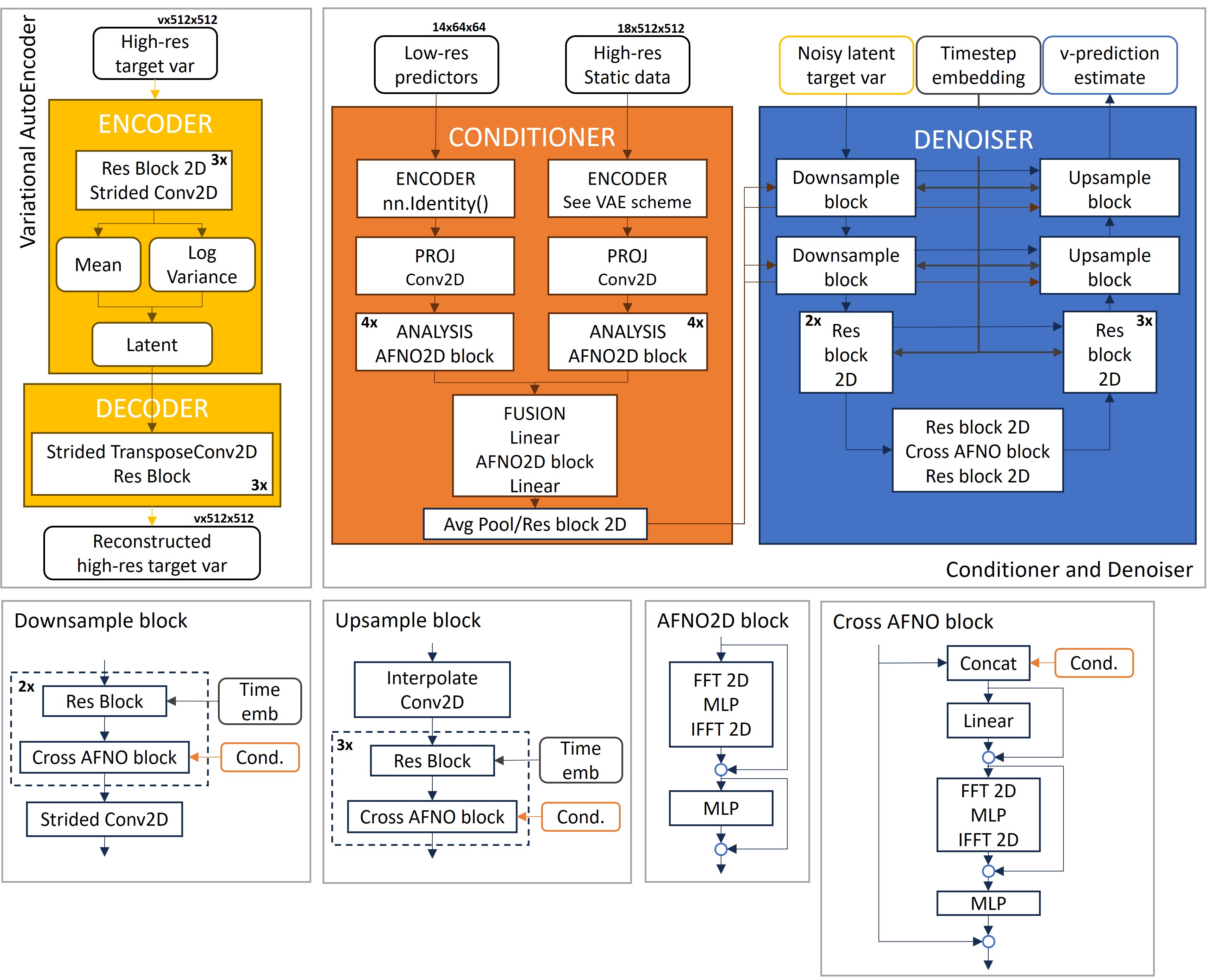}
    \caption{An overview of the components of our downscaling latent diffusion model: the Variational AutoEncoder, the Conditioner, and the Denoiser networks. \textit{Conv} denotes convolution. \textit{MLP} (multilayer perceptron) is a block consisting of a linear layer, activation function and another linear layer. \textit{Res block} denotes a ResNet-type residual block. \textit{v} in the array size labels stands for the number of target variables (1 for 2-m temperature and 2 for 10-m horizontal wind speed components).}
    \label{fig:LDM_components}
\end{figure}

For the downscaling task, the conditioning information consists of the low-resolution predictors' data and the high-resolution static data, elaborated by the conditioner. Figure \ref{fig:LDM_components} depicts details of the denoiser's structure.

To improve the reconstruction of extreme values (for both temperature and wind speed) we implemented the v-prediction parameterization in our LDM model, following \citet{DBLP:journals/corr/abs-2202-00512}: this parameterization trains the denoiser to model a weighted combination of both the noise and the start image, instead of either the only noise or the only start image as done in the more traditional implementations \textit{eps} and \textit{x0}, respectively.

\begin{figure}
    \centering
    \includegraphics[width=1\linewidth]{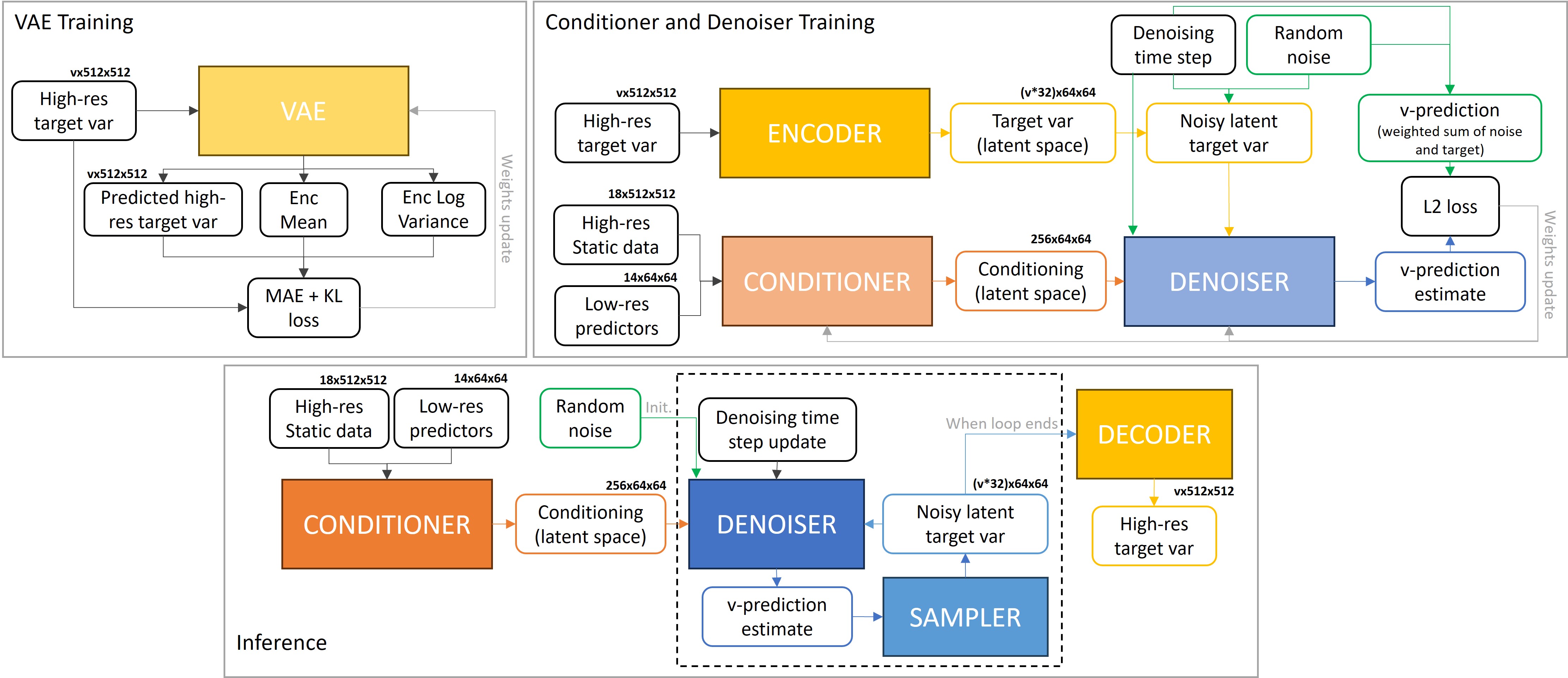}
    \caption{An overview of the training and inference procedures for our downscaling latent diffusion model. \textit{v} in the array size labels stands for the number of target variables (1 for 2-m temperature and 2 for 10-m horizontal wind speed components).}
    \label{fig:LDM_train_inf}
\end{figure}

As shown in Figure \ref{fig:LDM_train_inf}, the conditioner and the denoiser are trained together, minimizing the mean square error (L2), and feeding the network with random patches of ERA5 predictors (64x64 pixels) and static data (512x512 pixels) for the conditioning, and high-resolution target variables (512x512 pixels) for the ground truth. The training is performed using the AdamW optimizer \citep{loshchilov2018decoupled}, and Exponential Moving Averaging (EMA) is applied to the network weights, following \citet{StableDiffusion_2021}.

\section{Results}
\label{sec:results}
All the presented models were tested on a one-year dataset, which was held out during the training and validation processes. The results from the LDM\_res are evaluated based on a single inference run, obtained using 100 denoising steps; its potential to produce ensemble results is postponed to future analyses.

The following sections compare the results from LDM\_res against the baselines using various verification metrics and distributions. In the Supporting Information we report the comparison of results from the LDM trained with and without the residual approach to provide an overview of the improvements introduced by this method.

\subsection{Qualitative evaluation}
\label{subsec:QualitativeEval}
\begin{figure}
    \centering
    \includegraphics[width=0.58\linewidth]{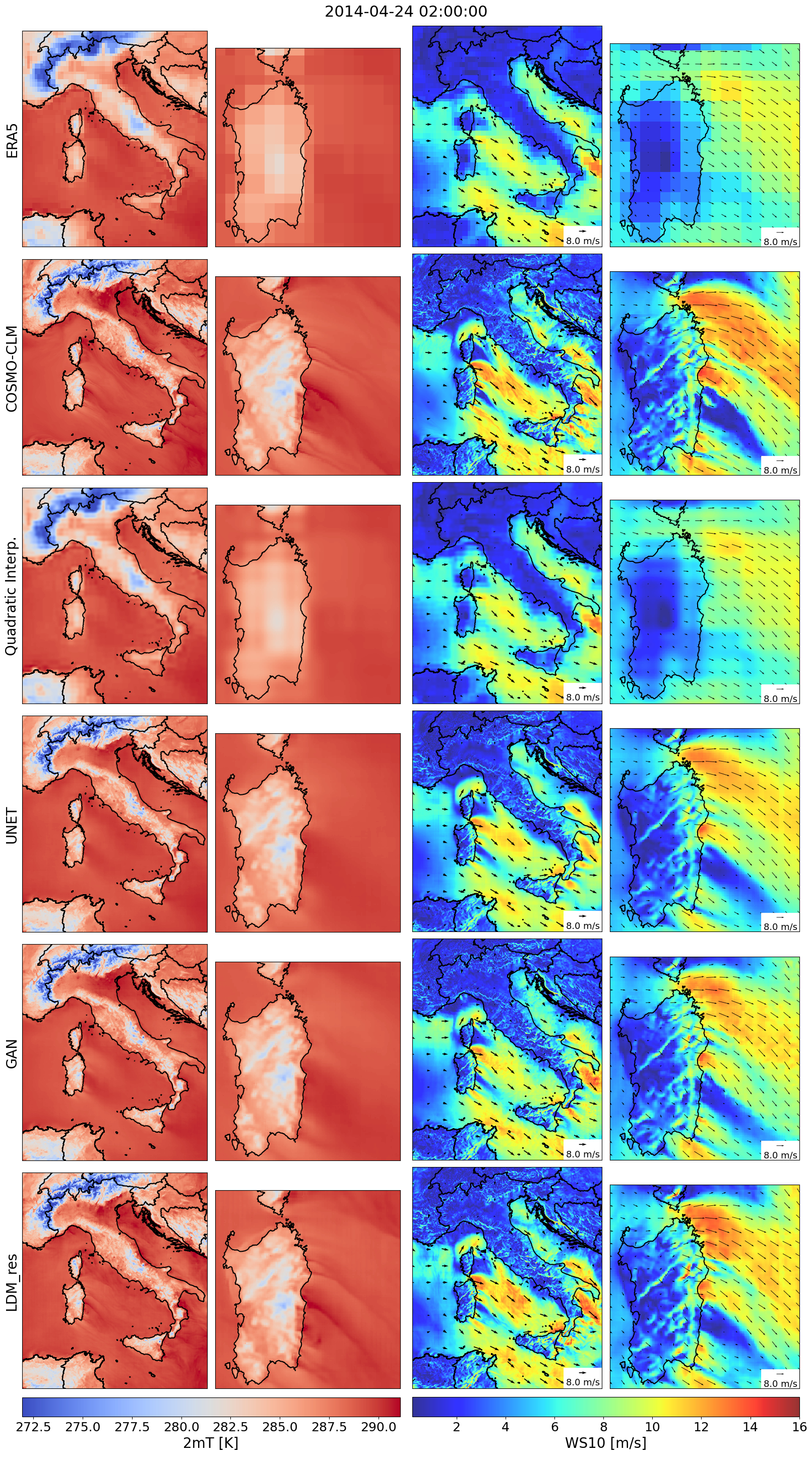}
    \caption{Downscaled variables from all the tested models against low-resolution ERA5 input data and high-resolution COSMO\_CLM reference truth, for a randomly picked timestamp. The left columns refer to 2-m temperature, and the right columns refer to 10-m wind speed. The second and fourth columns show a zoom-in on Sardinia Island.}
    \label{fig:snapshots}
\end{figure}

To provide a qualitative and perceptual overview of the obtained results, we present a random snapshot of downscaled variables compared with both the input ERA5 low-resolution data and the COSMO\_CLM high-resolution reference-truth (Figure \ref{fig:snapshots}). The second and third columns show a zoom-in on Sardinia Island, providing a deeper overview of models' performance over complex terrain, coastal shores, and open sea. Both generative models, the GAN and LDM\_res, effectively overcome the blurriness observed in both the quadratic interpolation and the UNET for the target variables. Particularly for 2-m temperatures, LDM\_res demonstrates a remarkable ability to identify and reconstruct discontinuities in the variable field (zoomed-in view in Figure \ref{fig:snapshots}).
Figure \ref{fig:snapshots} also includes results for 10-meter wind speed (in color), which is a derived field obtained by combining the two actual target variables of the models, U and V. Perceptually, the results for this variable from both the GAN and LDM\_res appear similar and equally plausible, displaying significantly more small-scale features compared to the UNET. A deeper qualitative examination reveals that the GAN aligns well with the reference truth, particularly over land, but exhibits mode collapse over the sea for both target variables. An example of this effect is shown in the Supporting Information. Conversely, the LDM\_res consistently generates plausible high-resolution data across the entire domain, over both land and sea, and for both target variables.

\subsection{Verification deterministic metrics}
\label{subsec:DetMetricsEval}
\begin{figure}
    \centering
    \includegraphics[width=0.85\linewidth]{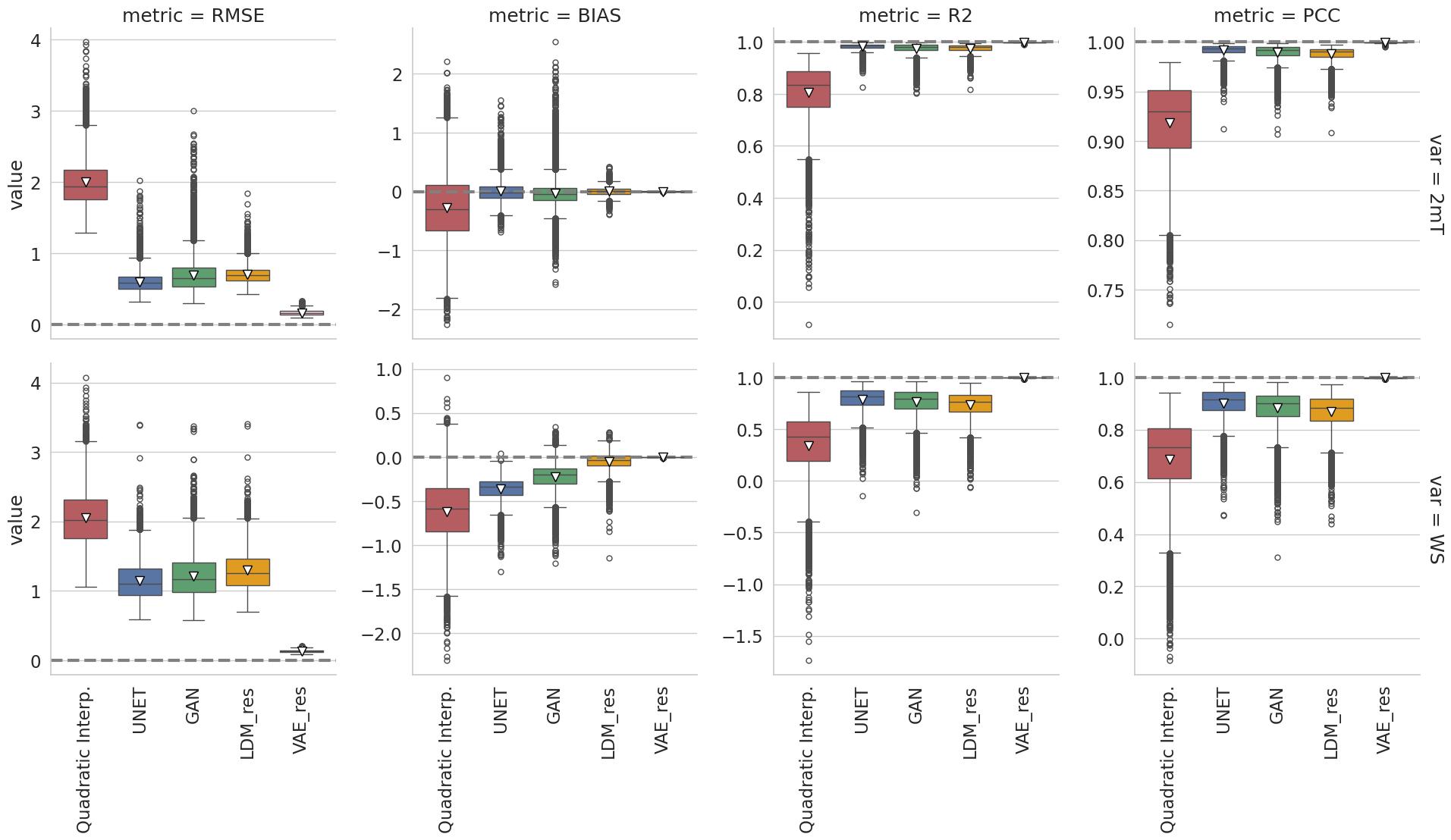}
    \caption{Comparison of deterministic metrics for spatially-averaged results of the analyzed models (top row refers to the 2-m temperature and bottom row refers to the 10-m wind speed). Notice that y axes are not shared between panels. The dashed line highlights the reference value for each metric, and the white triangle highlights the mean metric value.}
    \label{fig:deterministicMetrics}
\end{figure}

Figure \ref{fig:deterministicMetrics} compares model results for different deterministic metrics, averaging results over the whole domain for each test timestep. In addition to results from the baseline and tested models, Figure \ref{fig:deterministicMetrics} also reports results for the VAE of the LDM. These results are obtained using the VAE offline, feeding it with COSMO\_CLM high-resolution data and calculating the metrics on the reconstructed data: this allows quantifying the LDM error resulting from the data decompression from the latent space only. We present three distance metrics, the Root Mean Square Error (RMSE), the mean bias (BIAS), and the coefficient of determination (R2), and one correlation metric, the Pearson Correlation Coefficient (PCC) (all calculated following \citet{ray_bell_2021_5173153}).

The UNET and GAN models show comparable and best results for all metrics, except for the bias. Indeed, minimizing the MSE is the exact goal of their training procedure. Conversely, the LDM has been trained on a much different objective but performs very well for all the metrics. As expected, all models struggle more in downscaling the wind components than the 2-m temperature. Biases show that all models perform very well, with LDM\_res excelling, especially for temperature. The UNET and the GAN show spatially averaged biases within 1\textdegree C for temperature, while LDM\_res shrinks this variability to less than 0.5\textdegree C. Spatially averaged biases amount to 1 m/s for wind speed, with a narrower spread for LDM\_res. The UNET and LDM\_res show a less skewed distribution than the GAN model for the 2-m temperature: while the GAN model tends to underestimate the average 2-m temperature mostly, LDM\_res shows a very balanced distribution for over- and under-estimations. As for the wind speed biases, all the models always slightly underestimate the target variable.

The results show that the VAE contribution to LDM\_res is the highest for the RMSE of 2-m temperature, while bias, R2, and PCC have very little to no effects on temperature and wind speed. 

\subsection{Spatial distribution of errors}
\label{subsec:spatdistrib}

\begin{figure}
    \centering
    \includegraphics[width=0.9\linewidth]{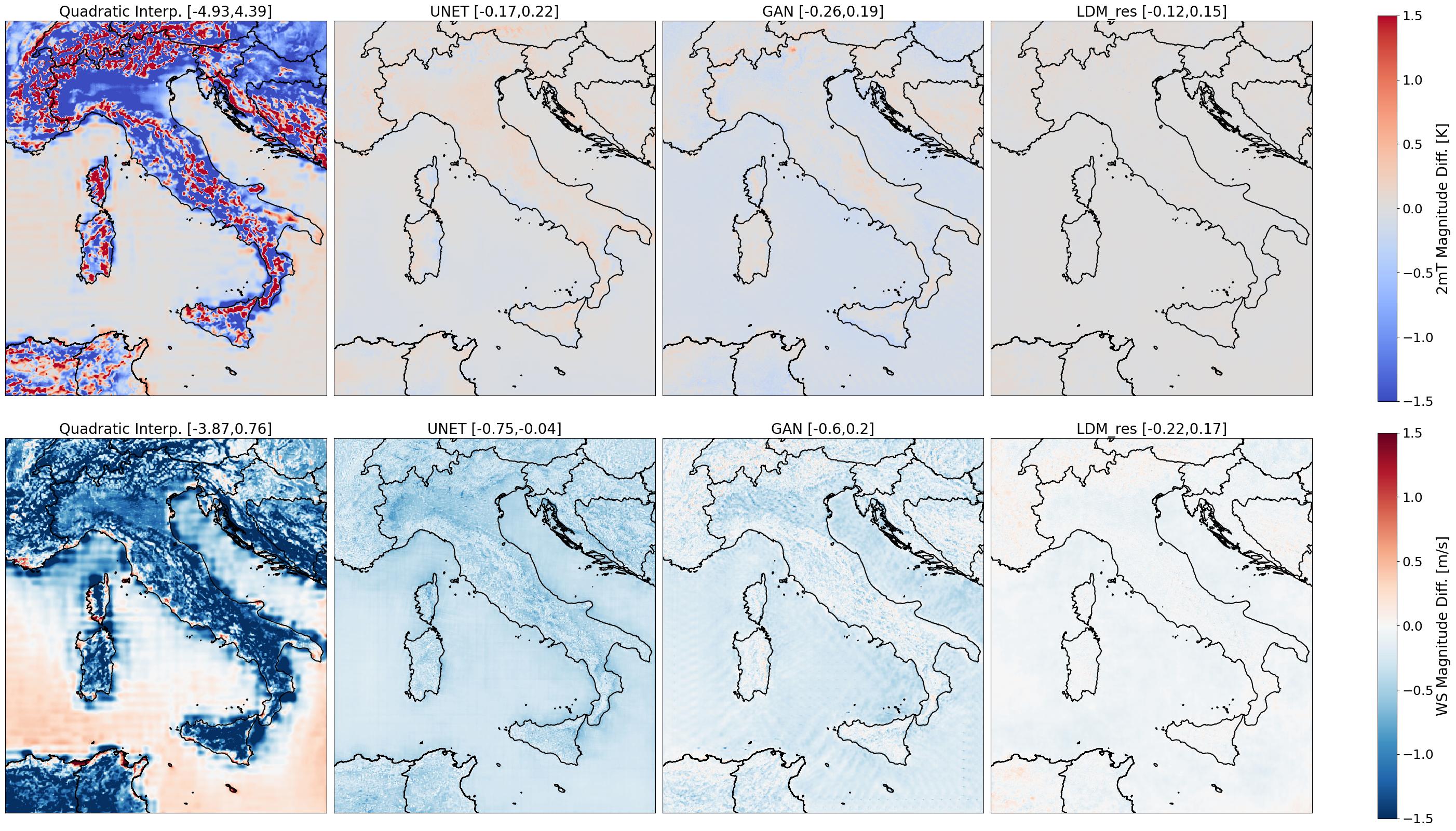}
    \caption{Spatial distribution of averaged-in-time magnitude difference for each tested model. The top row refers to the 2-m temperature, and the bottom row refers to the 10-m wind speed.
    }
    \label{fig:spat_errors}
\end{figure}

The spatial distribution of averaged-in-time magnitude differences for both the target variables and all tested models is illustrated in Figure \ref{fig:spat_errors}. Within each panel, the numbers in squared brackets represent the 0.5 and 99.5 percentile values, offering insight into the highest errors recorded over the domain. Negative and positive values signify underestimation and overestimation, respectively, for both variables. Results from the quadratic interpolation of ERA5 data provide information on the original input data: 2-m temperature tends to be highly overestimated over complex terrain but underestimated on flat terrain, with smaller errors over the sea. Wind speed, conversely, is largely underestimated over land, particularly over mountain ridges, with a tendency towards overestimation along coastal shores. 

On the contrary, all DL-based models, including the UNET baseline, exhibit substantially smaller errors. For 2-m temperature, errors remain below 0.3°C, while for wind speed, they stay under 0.8 m/s across the entire domain.
Notably, the UNET and GAN models perform comparably well for 2-m temperature, whereas LDM\_res excels, leveraging diffusion processes to reduce the UNET errors homogeneously.

As for the wind speed results, all models exhibit a tendency to underestimation. LDM\_res demonstrates superior performance, minimizing errors to nearly zero over most of the domain, with a uniform distribution over land and sea. The GAN displays traces of its characteristic mode collapses, especially over the sea: this evidence indicates that these mode collapses persist statically in fixed locations over time, consistently with the deployed training approach (i.e. feeding the network always the entire, fixed domain).

\subsection{Frequency distributions}
\label{subsec:FreqDistr} 
\begin{figure}
    \centering
    \includegraphics[width=0.84\linewidth]{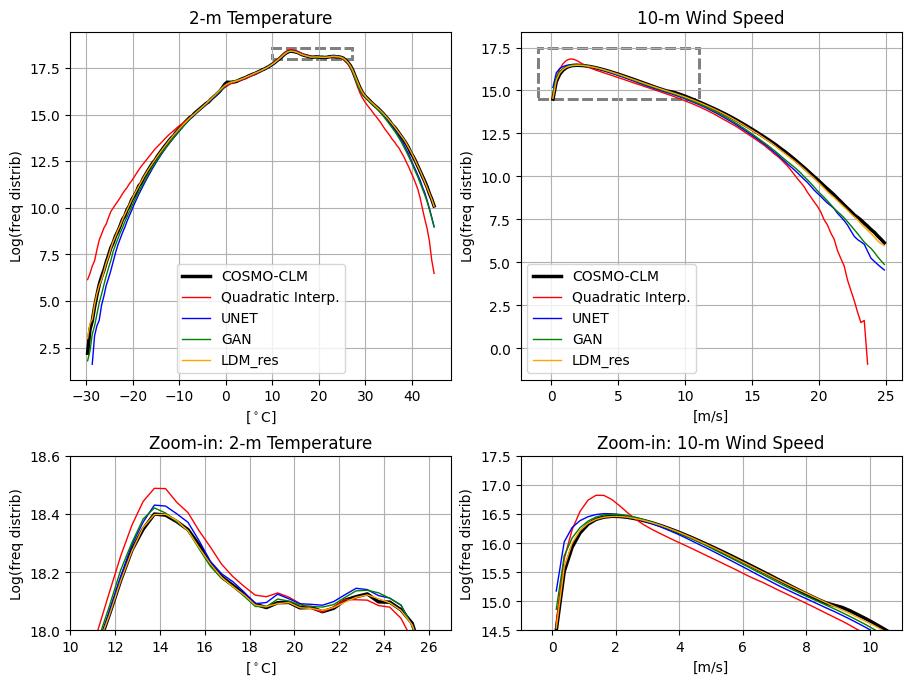}
    \caption{Comparison of frequency distributions for results from the tested models against COSMO\_CLM reference-truth. The left column refers to the 2-m temperature, and the right column refers to the 10-m wind speed. Counting of pixel-wise data is cumulated for the yearly test dataset over bins of 0.5 \textdegree C and 0.05 m/s for temperature and wind speed, respectively. Notice that y-axes are logarithmic to highlight the tails of the distributions, hence the extreme values. The top row focuses on the tails of the distributions, i.e. on extreme values, while the bottom row focuses on the most frequent values and is a zoom-in on the dashed boxes for each variable. }
    \label{fig:FreqDistr}
\end{figure}

Figure \ref{fig:FreqDistr} presents the results in terms of frequency distributions. LDM\_res precisely captures the reconstruction of the 2-m temperature frequency distribution, surpassing all other models. All DL models effectively mitigate the occurrence of cold extremes evident in the low-resolution data (as demonstrated by the quadratic interpolation distribution) while increasing the incidence of warm extremes.
Notably, the adversarial training of the UNET yields marginal enhancements in capturing the frequency distribution, with the GAN slightly outperforming the UNET, particularly regarding cold extremes.
Conversely, the diffusion process performed by LDM\_res significantly corrects the UNET residual errors, aligning closely with the reference-truth distribution across all temperature values.

Reconstructing the distribution of 10-m wind speed proves more challenging for all models, given the inherent chaotic nature of the $U$ and $V$ wind components compared to temperature, which is strongly influenced by terrain elevation. Nonetheless, performance outcomes mirror those of the 2-m temperature. The GAN modestly improves upon UNET results, primarily in reducing occurrences of low wind speeds. LDM\_res exhibits the highest performance in capturing both the tail and center of the wind speed distribution.

\subsection{Radially Averaged Power Spectral Density (RAPSD)}
\label{subsec:RAPSD} 

\begin{figure}
    \centering
    \includegraphics[width=0.84\linewidth]{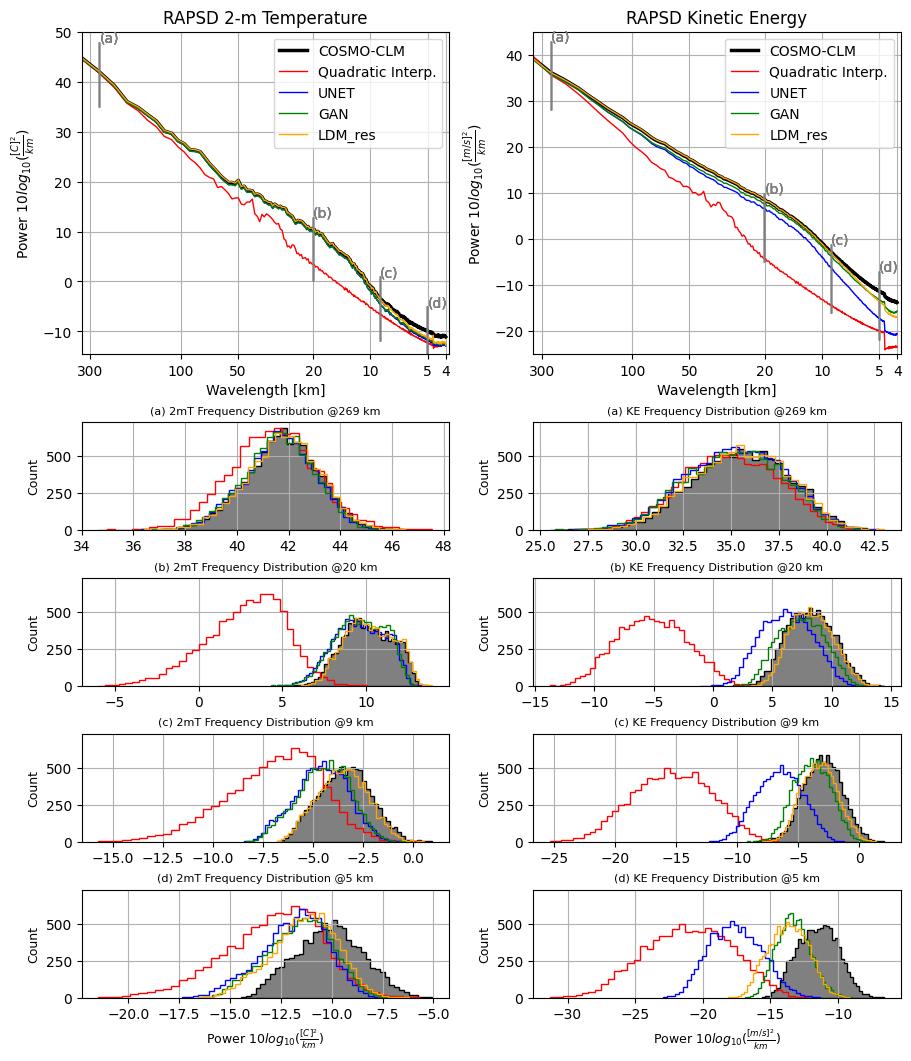}
    \caption{Comparison of Radially Averaged Power Spectral Density (RAPSD) distributions for results from the tested models against COSMO\_CLM reference-truth. The left column refers to the 2-m temperature, and the right column refers to the 10-m wind speed. The first row shows the averaged-in-time spectra, across the whole test dataset. Notice that in the first row y-axes are logarithmic to highlight the tail of the distributions, hence the high frequencies. The bottom rows show the distributions of single-time RAPSD values for fixed wavelengths, namely 269, 20, 9, and 5 km. }
    \label{fig:SpectraDistrib}
\end{figure}

Figure \ref{fig:SpectraDistrib} showcases the results in terms of Radially Averaged Power Spectral Density (RAPSD), computed following the implementation outlined in \citet{pysteps}. The top row of the figure illustrates a single RAPSD, representing the average of each RAPSD calculated for every timestamp within the test dataset. To provide insight into the distribution of these values across all timestamps, the distributions of single-time RAPSD for fixed wavelengths are displayed in the bottom rows of Figure \ref{fig:SpectraDistrib}.

Overall, all DL models effectively reconstruct the 2-m temperature power spectra down to wavelengths of 10 km. However, LDM\_res consistently outperforms both the UNET and the GAN, as evident in panels (b) and (c) of Figure \ref{fig:SpectraDistrib}.  The UNET and the GAN yield similar results, with marginal yet consistent enhancements originating from the adversarial training of the UNET. The diffusion process of LDM\_res adeptly enhances the generation of small-scale features, showing precise reconstruction of the spectra up to 9 km (see panel c). For scales smaller than 9-10 km, all models exhibit decreased performance, albeit still showing improvements over the quadratic interpolation of ERA5. LDM\_res still outperforms the other models but the very small-scale variability of the original data is slightly underestimated. This behavior is to be ascribed to the VAE, as further elucidated in the Supporting Information. Indeed, original reference-truth data compressed and reconstructed by the VAE show the very same power spectra underestimation for scales smaller than 9 km. The loss of information is therefore due to and inherent to the projection to the latent space.    

In contrast, results for wind speed distinctly demonstrate that generative models surpass both quadratic interpolation and UNET, effectively matching the slope of the energy power spectra and remaining competitive with each other. Specifically, LDM\_res consistently outperforms the GAN up to 7 km, as emphasized in panels (b) and (c) of Figure \ref{fig:SpectraDistrib} (right column). Similar to the 2-m temperature, for scales smaller than 10-9 km, both the GAN and LDM\_res experience reduced performance, although they consistently exhibit improvements over the UNET. This behavior, for LDM\_res, is in this case only partly to be ascribed to the VAE, as further shown in the Supporting Information, and an additional loss, for scales smaller than 9 km, is to be attributed intrinsically to the extraction of features with the diffusion process conditioned with the low-resolution data and high-resolution static data.

\subsection{Run-time performance}
\label{subsec:runtime_performances}
In this section, we compare the run-time performance of our tested models. These characteristics are of fundamental importance given the potential target applications of such models. Table \ref{tab:run_times} reports data for each model: to give a whole picture of the needed resources we provide information on both the training and inference requirements. The hour budgets indicated for LDM\_res include hours to train/run the UNET, which is needed to provide the residual data, and the VAE\_res. LDM\_res budgets are therefore comprehensive of the required computational time to train and run the whole modeling chain from scratch. Simulations were run either on NVIDIA GeForce RTX 4090 or NVIDIA A100 GPUs. The training dataset comprises ~129000 hourly samples over a target domain of 576x672 pixels (at high resolution) and 72x86 pixels (at low resolution). The UNET and GAN training ran with 4-size batches for both target variables on the whole target domain. LDM\_res training ran with 8- and 4-size batches for the two target variables (2-m temperature and 10-m wind components, respectively) on patches of 512x512 and 64x64 for high-resolution and low-resolution data, respectively. Inference times are calculated running with single-dimension batches, on the size of the entire domain. 

As shown in Table \ref{tab:run_times}, LDM\_res implies more expensive training and inference processes when compared with the tested DL baselines. This evidence is expected, given the more complex structure of the diffusion model and its dimensions in terms of trainable parameters, which is an order of magnitude greater than that of the baselines. Nevertheless, the required computational time for both training and inference remains contained and competitive with the other available options. LDM\_res requires ~10 days over 8 GPUs to train the models for both the target variables and ~30 hours on a single GPU to downscale one year of hourly temperature and wind data. As a term of comparison, we note here that a 1-year-long COSMO\_CLM simulation ran over the very same domain requires 61 h, running on 2160 cores \citep{Raffa2021} (of course producing many more high-resolution variables than the sole 2-m temperature and 10-m horizontal wind components).

\begin{table}
 \caption{Number of GPU hours required for the training and inference (of a 1-year-long test set) by each tested model.}
  \centering
  \begin{tabular}{lcccc}
    \hline
    \multirow{2}{*}{\textbf{Model}} &\multicolumn{2}{c}{\textbf{Training}} & \multicolumn{2}{c}{\textbf{Inference}}\\
    & \textbf{2mT}& \textbf{UV}& \textbf{2mT}& \textbf{UV} \\
    \hline
    UNET     &   $\sim$250    & $\sim$380 & $\sim$1 & $\sim$1\\  
    GAN     &   $\sim$300    & $\sim$100 & $\sim$1 & $\sim$1\\   
    LDM\_res     &   $\sim$870    & $\sim$1100 & $\sim$15 & $\sim$16\\   
    \hline
  \end{tabular}
  \label{tab:run_times}
\end{table}

\section{Discussion and Conclusions}
\label{sec:discussion}

This study compares the performance of various downscaling models, focusing on their ability to reconstruct high-resolution meteorological variables from low-resolution input data. The models evaluated include a baseline UNET, a Generative Adversarial Network (GAN), and a Latent Diffusion Model with a residual approach against the reference UNET (LDM\_res). The results are analyzed using qualitative evaluations, deterministic metrics, spatial error distributions, frequency distributions, Radially Averaged Power Spectral Density (RAPSD), and runtime performance.

LDM\_res demonstrates superior performance across most metrics, particularly in reconstructing fine-scale details and maintaining accuracy in frequency distributions (especially for the extreme values) and spatial error distributions. LDM\_res outperforms the other models in reconstructing the power spectra, showing superior performance, especially for wind speed, with outstanding results up to 7 km wavelengths. Residual errors at smaller scales can be attributed to the data projection into the latent space, specifically to the usage of the VAE. This performance loss might be mitigated by conducting the diffusion process directly in the pixel space. However, this alternative approach would substantially increase the computational costs for both training and inference.

The remarkable results of LDM\_res are to be ascribed equally to two fundamental aspects of the proposed model:
\begin{itemize}
\item the incomparable effectiveness of the diffusion process in extracting features and leveraging the provided conditioning;
\item the residual approach which allows the diffusion process to focus only on smaller scales and more subtle characteristics of the fields, delegating the estimates of large-scale variation of the atmospheric fields to a simpler, yet effective, network.
\end{itemize}

However, the great performance of LDM\_res comes at the cost of significantly higher computational requirements for both training and inference, when compared to the other DL models, the UNET or the GAN. Nonetheless, LDM\_res still offers a significant advantage in terms of inference speed and computational efficiency once the model is trained when compared to the extensive computational resources required by COSMO\_CLM.

In conclusion, the ability of LDM\_res to accurately reproduce the statistics of the COSMO\_CLM model reference truth data, provided with the same input, demonstrates its potential as an effective and versatile dynamical downscaling emulator. This approach significantly accelerates the downscaling process compared to traditional numerical dynamical models, making it highly suitable for a broad range of important applications, such as downscaling seasonal forecasts or climate projections. However, a primary limitation of such deep learning models remains the temporal consistency of the generated fields which is not provided by the construction of the models.

\section{Future work}
\label{sec:futurework}
The results presented in this work suggest several promising directions for further investigation into the application of latent diffusion models for downscaling. Future research could explore the ensemble generation capabilities of our Latent Diffusion Model, its effectiveness in downscaling discontinuous and chaotic variables such as precipitation, crucial for many applications, and the temporal consistency of the downscaled data, including methods to enforce this consistency within the model architecture. Long-term developments may include applying latent diffusion models to real-time weather forecasts, seasonal forecasts, and climate projections, with adjustments in the training procedure, particularly in selecting low-resolution input predictors and target reference truths.


\codedataavailability{Data from ERA5 \citep{Hersbach2020} was used as input low-resolution data for our models and was downloaded from the Copernicus Climate Change Service \citep{ERA5pressure, ERA5surface}. The results contain modified Copernicus Climate Change Service information 2020. Neither the European Commission nor ECMWF is responsible for any use that may be made of the Copernicus information or data it contains. The target high-resolution VHR-REA\_IT data \citep{Raffa2021, Adinolfi2023} was downloaded from the Euro-Mediterranean Center on Climate Change (CMCC) Data Delivery System \citep{VHRREAIT}. CMCC produced the VHR-REA\_IT dataset as a part of the Highlander project and released it under the CC BY 4.0 Licence. The full, preprocessed dataset used for the presented experiments is available on Zenodo \citep{tomasi_2024_12944960, tomasi_2024_12945014, tomasi_2024_12945028, tomasi_2024_12945040, tomasi_2024_12945050, tomasi_2024_12945058, tomasi_2024_12945066}. Additionally, Zenodo hosts a sample dataset \citep{tomasi_2024_12934521} to test and train the models and the trained models themselves \citep{tomasi_2024_12941117}. A dedicated GitHub repository (https://github.com/DSIP-FBK/DiffScaler) hosts the Pytorch Lightning \citep{Falcon_PyTorch_Lightning_2019} code of the models described in this paper, based on the Lightning-Hydra-Template \citep{Yadan2019Hydra}, licensed under the MIT License. The repository also hosts the code to reproduce all the images shown in this paper. LDM\_res v1.0 GitHub release is archived on Zenodo \citep{zenodo_ldm_res_repo} and allows to download the code to reproduce the presented experiments.} 

\appendix
\section{Additional snapshots of downscaled data}
\label{subsec:QualitativeEval_extra}

Some additional snapshots of downscaled data from all the tested models are shown in the following Figures \ref{fig:snapshots_extra} and \ref{fig:snapshots_extra2}. The left columns refer to 2-m temperature, and the right columns refer to 10-m wind speed. The second and fourth columns show a zoom-in on Sardinia Island.

\begin{figure}
    \centering
    \noindent\includegraphics[width=0.58\linewidth]{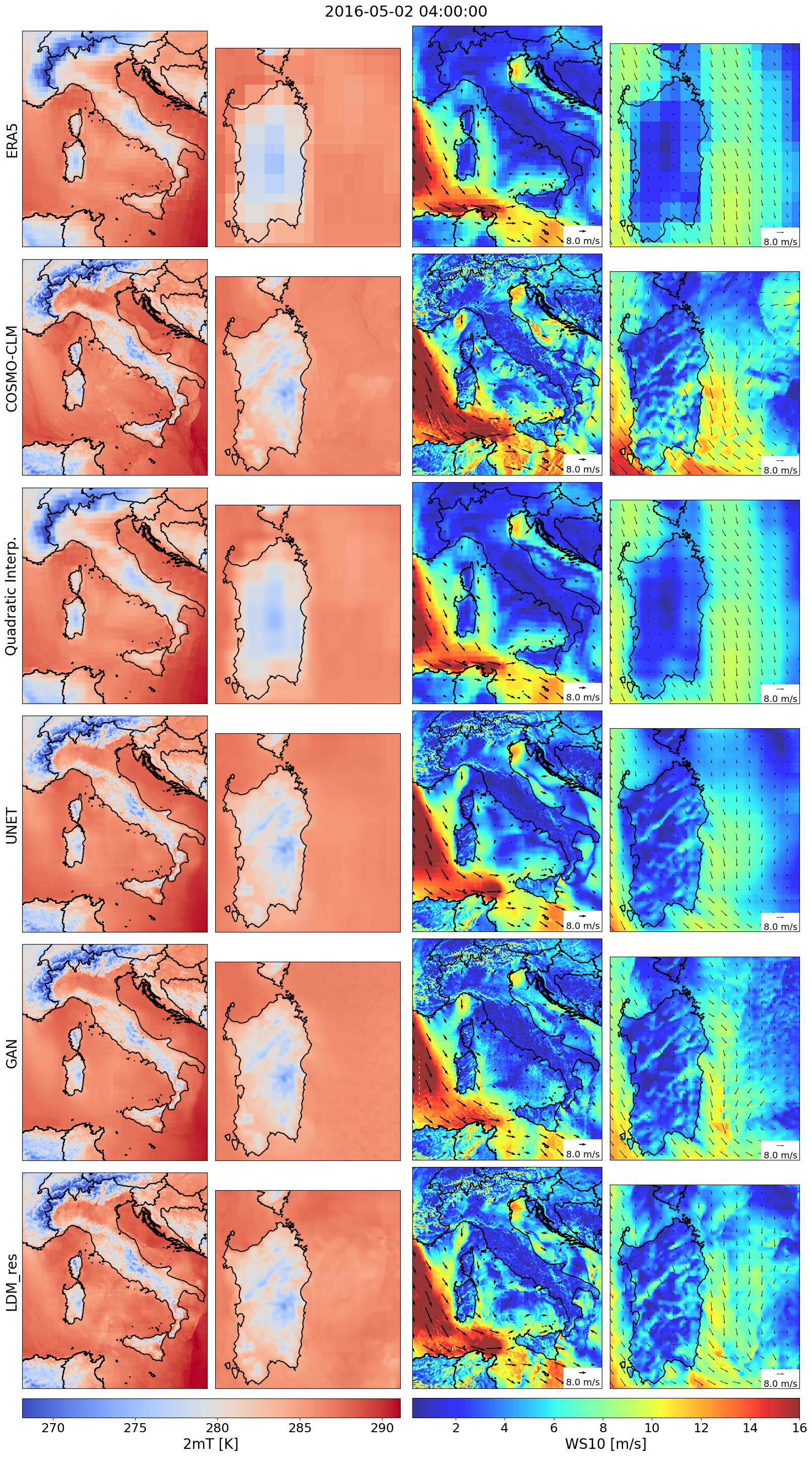}
    \caption{Downscaled variables from all the tested models against low-resolution ERA5 input data and high-resolution COSMO-CLM reference truth, for an additional random timestamp.}
    \label{fig:snapshots_extra}
\end{figure}

\begin{figure}
    \centering
    \noindent\includegraphics[width=0.58\linewidth]{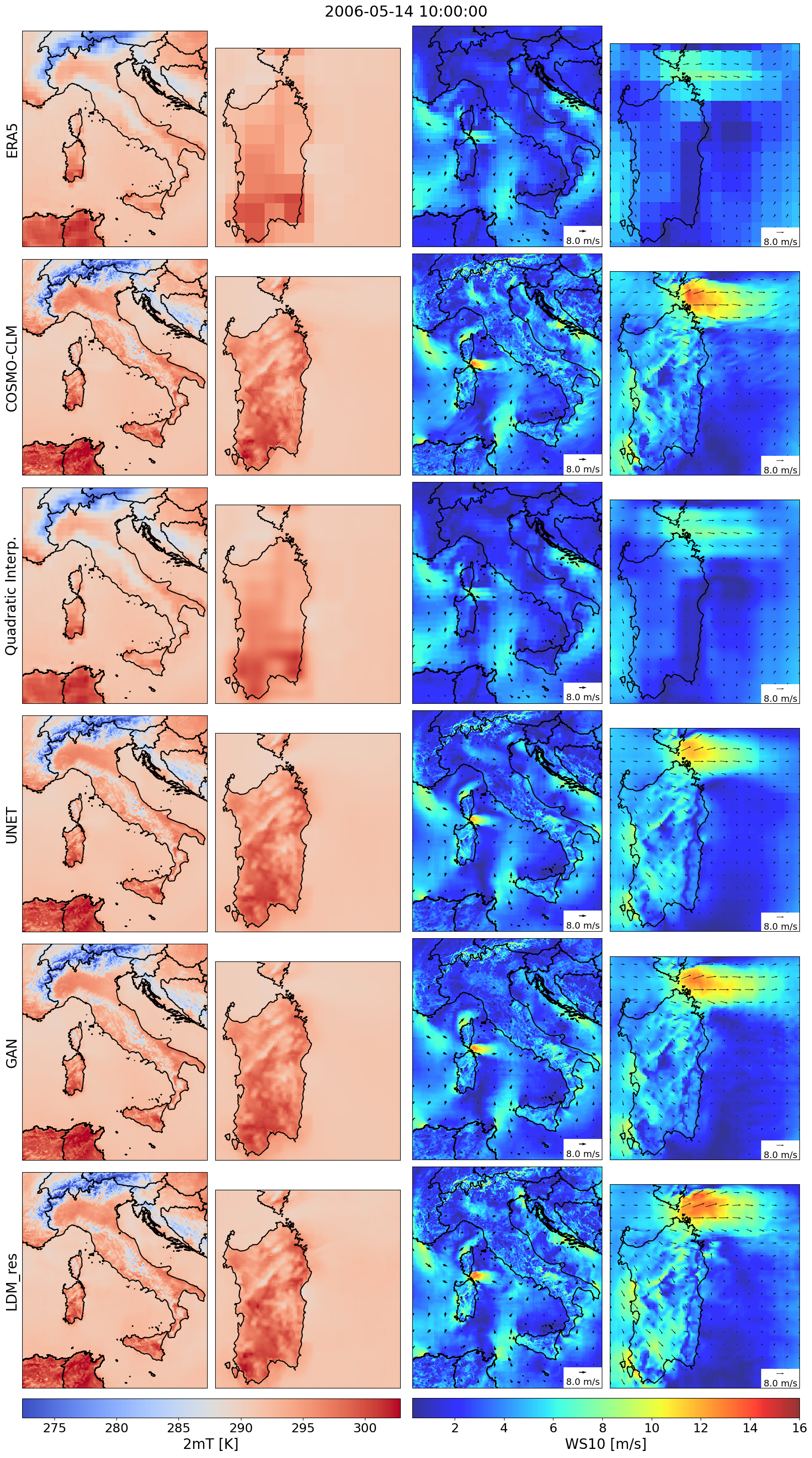}
    \caption{Downscaled variables from all the tested models against low-resolution ERA5 input data and high-resolution COSMO-CLM reference truth, for an additional random timestamp.}
    \label{fig:snapshots_extra2}
\end{figure}

\section{Calculation of verification metrics}
\label{subsec:DetMetricsEval_extra}
In Section \ref{subsec:DetMetricsEval}, we discuss the following metrics: three distance metrics, the Root Mean Square Error (RMSE), the mean bias (BIAS), the coefficient of determination (R2), and one correlation metric, the Pearson Correlation Coefficient (PCC). All the metrics are calculated with the xskillscore library \cite{ray_bell_2021_5173153}. Definitions are the following:

\begin{equation}
\mathrm{RMSE} = \sqrt{\frac{1}{n}\sum_{i=1}^{n}(a_{i} - b_{i})^{2}}
\end{equation}

\begin{equation}
\mathrm{BIAS} = \frac{1}{n}\sum_{i=1}^{n}(a_{i} - b_{i})
\end{equation}

\begin{equation}
SS_{{tot}} = \sum_{i=1}^{n} (a_i - \bar{a})^2
\end{equation}

\begin{equation}
SS_{{res}} = \sum_{i=1}^{n} (a_i - b_i)^2
\end{equation}

\begin{equation}
R^2 = 1 - \frac{SS_{{res}}}{SS_{{tot}}}
\end{equation}

\begin{equation}
PCC = \frac{ \sum_{i=1}^{n} (a_{i} - \bar{a}) (b_{i} - \bar{b}) } {\sqrt{ \sum_{i=1}^{n} (a_{i} - \bar{a})^{2} } \sqrt{ \sum_{i=1}^{n} (b_{i} - \bar{b})^{2} }}
\end{equation}

where $a$ and $b$ are the predicted and reference truth values, respectively, for every $i^{th}$ pixel of the domain; $n$ are the total number of pixels in the domain.

\section{On the contribution of the VAE}     
\label{subsec:VAEcontrib}
\begin{figure}
    \centering
    \noindent\includegraphics[width=0.79\linewidth]{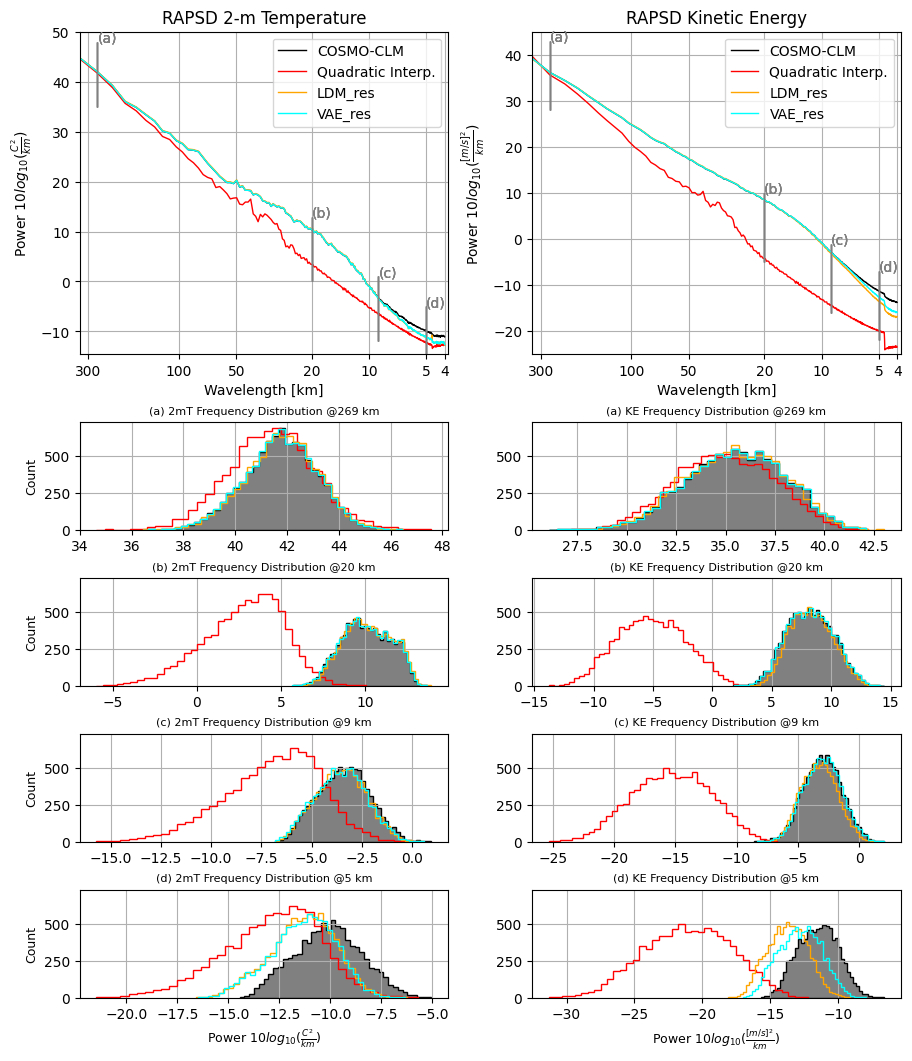}
    \caption{Comparison of Radially Averaged Power Spectral Density (RAPSD) distributions for results from LDM\_res and  VAE\_res against COSMO-CLM reference-truth and ERA5 quadratic interpolation. The left column refers to the 2-m temperature, and the right column refers to the 10-m wind speed. The first row shows the averaged-in-time spectra, across the whole test dataset. Notice that in the first row y-axes are logarithmic to highlight the tail of the distributions, hence the high frequencies. The bottom rows show the distributions of single-time RAPSD values for fixed wavelengths, namely 269, 20, 9, and 5 km. }
    \label{fig:SpectraDistrib_VAE}
\end{figure}

Figure \ref{fig:SpectraDistrib_VAE} compares power spectra from the COSMO-CLM test reference truth data with power spectra from high-resolution test data generated using three models: the quadratic interpolation of ERA5 data, VAE\_res, and LDM\_res. This figure highlights the decompression stage's contribution from the latent space to the pixel space in LDM\_res.

VAE\_res takes as input the original reference truth high-resolution target variables. LDM\_res takes as input the low-resolution ERA5 predictors, the high-resolution static data, and random noise, produces information in the latent space, and projects them in the pixel space using the VAE Decoder. Both VAE\_res and LDM\_res also use the corresponding UNET estimates for each target variable, but these quantities are subtracted before the encoding step (when applied) and added back after the decompression stage, acting essentially as constants. Therefore, the reconstruction errors in the power spectra for VAE\_res are thus solely attributed to the compression/decompression processes, while the reconstruction errors in the power spectra for LDM\_res arise from both the diffusion and decompression processes.

For 2-m temperature, the power spectra from the VAE\_res and LDM\_res are nearly identical across all wavelengths, including the smallest scales. This indicates that the errors in reconstructing the spectra with LDM\_res are attributable solely to the decompression stage, with the diffusion process effectively and accurately extracting latent features from the conditional data. On the contrary, for 10-m wind speed, a more chaotic field, discrepancies between the power spectra from LDM\_res and VAE\_res are observed at scales smaller than approximately 7 km. These differences highlight the errors introduced by the sole extraction of features by the diffusion process.

\section{LDM residual versus non-residual results}
\label{subsec:res_vs_no_res}

\begin{figure}
    \centering
    \noindent\includegraphics[width=0.85\linewidth]{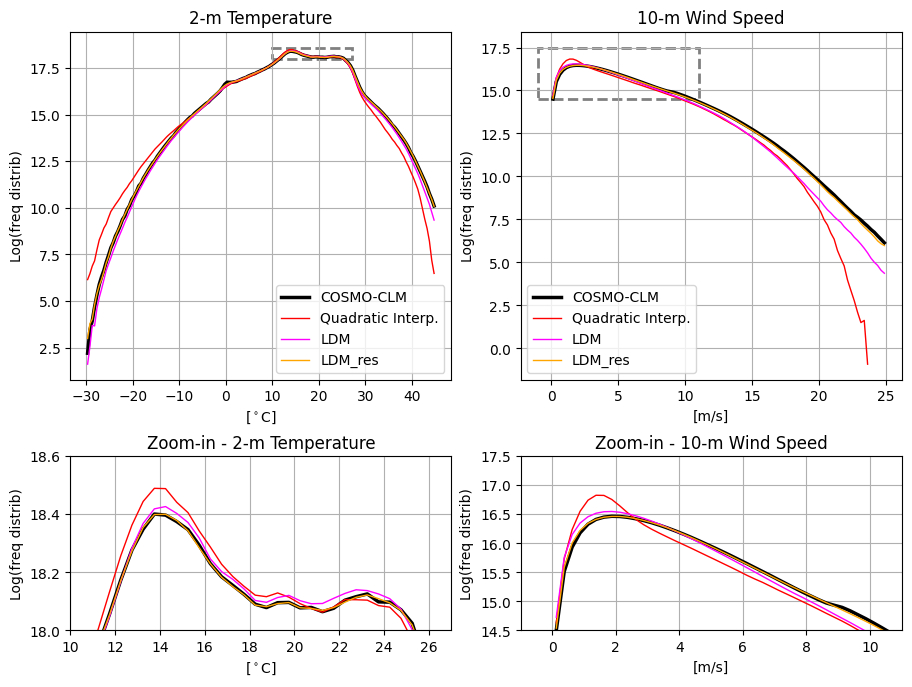}
    \caption{Comparison of frequency distributions for results from LDM and LDM\_res against COSMO-CLM reference-truth and ERA5 quadratic interpolation. The left column refers to the 2-m temperature, and the right column refers to the 10-m wind speed. Counting of pixel-wise data is cumulated for the yearly test dataset over bins of 0.5 \textdegree C and 0.05 m/s for temperature and wind speed, respectively. Notice that y-axes are logarithmic to highlight the tails of the distributions, hence the extreme values. The top row focuses on the tails of the distributions, i.e. on extreme values, while the bottom row focuses on the most frequent values and is a zoom-in on the dashed boxes for each variable. }
    \label{fig:FreqDistr_res_vs_nores}
\end{figure}

\begin{figure}
    \centering
    \noindent\includegraphics[width=0.79\linewidth]{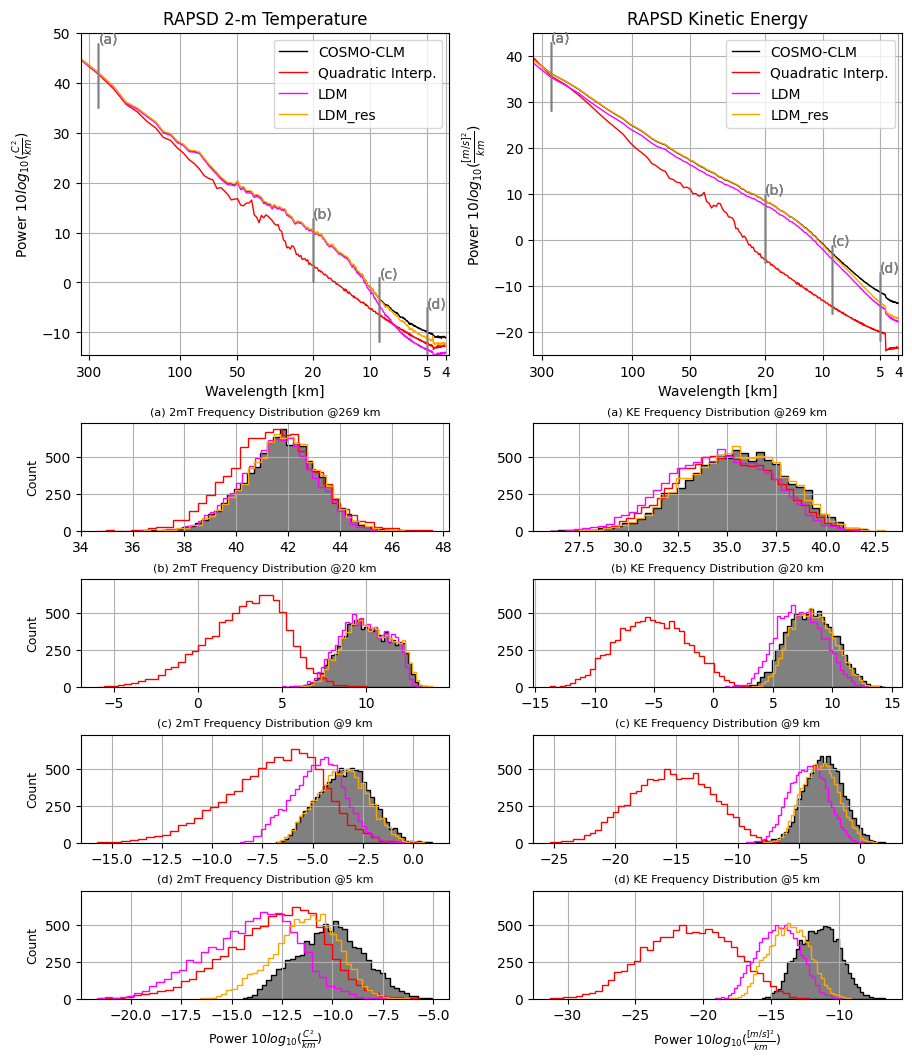}
    \caption{Comparison of Radially Averaged Power Spectral Density (RAPSD) distributions for results from LDM and LDM\_res against COSMO-CLM reference-truth and ERA5 quadratic interpolation. The left column refers to the 2-m temperature, and the right column refers to the 10-m wind speed. The first row shows the averaged-in-time spectra, across the whole test dataset. Notice that in the first row y-axes are logarithmic to highlight the tail of the distributions, hence the high frequencies. The bottom rows show the distributions of single-time RAPSD values for fixed wavelengths, namely 269, 20, 9, and 5 km. }
    \label{fig:SpectraDistrib_res_vs_nores}
\end{figure}

 In this section, we report the comparison of results from the LDM trained with and without the residual approach, to provide an overview of the improvements introduced by this approach. We compare results in terms of frequency distribution and of Radially Averaged Power Spectral Density, shown in Figures \ref{fig:FreqDistr_res_vs_nores} and \ref{fig:SpectraDistrib_res_vs_nores}.

 The graphs show relevant differences in the performance of the two models, especially in: (i) the estimation of the most frequent values of 2-m temperature, (ii) the reconstruction of the whole frequency distribution of the wind speed, (iii) the reconstruction of the 2-m temperature power spectra at small scales, with performances of LDM with no residual approach degrading below the quadratic interpolation of ERA5, and (iv) the reconstruction of the 10-m wind speed power spectra at all scales, with a quasi-constant lag between the two models throughout all wavelengths.
 
 The corresponding VAEs for the two models (VAE and VAE\_res) perform the same except for the small scales of the 2-m temperature power spectra (not shown). Thus, the lack of performance of LDM can be attributed to the VAE only for this occasion (i.e., (iii)). All the other listed deficiencies are to be solely attributed to the diffusion process itself, which, when trained to reconstruct a residual field instead of the original target variable field, gains performance in the reconstruction of all the frequency distributions, and of the power spectra across all wavelengths (not only the small scales), especially for chaotic variables such as wind speed.

\noappendix       




\appendixfigures  



\authorcontribution{All authors conceived and conceptualized the study. ET and GF designed the LDM\_res architecture and implemented the code for all the tested models. ET managed the data, ran the experiments, performed the analysis and verification of the results, and wrote the manuscript. All authors revised the results and reviewed the manuscript. MC supervised the study from end to end.} 

\competinginterests{The authors declare that they have no conflict of interest.} 


\begin{acknowledgements}
This work was developed with financial support from ICSC–Centro Nazionale di Ricerca in High Performance Computing, Big Data and Quantum Computing, Spoke4 - Earth and Climate, funded by European Union – NextGenerationEU.
\end{acknowledgements}







\bibliographystyle{copernicus}
\bibliography{references.bib}

\end{document}